%% file: main.tex
\documentclass[10pt,twocolumn,letterpaper]{article}
\usepackage[pagenumbers]{cvpr}
\usepackage{colortbl}
\definecolor{prior}{HTML}{4F6A8F}
\definecolor{priorbg}{HTML}{E9EEF5}
\definecolor{ours}{HTML}{B03A2E}
\definecolor{oursbg}{HTML}{F8E6E0}
\definecolor{reality}{HTML}{3A3A3A}
\definecolor{failure}{HTML}{D99A2B}
\usepackage{pifont}
\newcommand{\yes}{\ding{51}}
\newcommand{\no}{\ding{55}}

\newcommand{\oursb}[1]{\textbf{\textcolor{ours}{#1}}}
\usepackage{tabularx}
\usepackage{multirow}
\definecolor{tableblue}{HTML}{647F9E}
\usepackage{CJKutf8}
\definecolor{cvprblue}{rgb}{0.21,0.49,0.74}
\usepackage[breaklinks,colorlinks,allcolors=cvprblue]{hyperref}
\usepackage{etoolbox}

\makeatletter
\patchcmd{\@maketitle}{\vskip .375in}{\vspace*{-20pt}}{}{\PackageWarning{title}{Patch failed}}
\patchcmd{\@maketitle}{\vspace*{24pt}}{\vspace*{16pt}}{}{}
\makeatother

\title{Most of the Way There: Turning a Strong Video Prior into an Action-Conditioned World Model}

\author{
Zanyi Wang$^{1,2}$ \quad Xia Chen$^{2}$ \quad Liu Liu$^{2,\dagger}$ \quad Xianglong Guo$^{3}$ \quad Dengyang Jiang$^{4}$\\
Haodong Li$^{1}$ \quad Tingyang Xiao$^{2}$ \quad Wenhao Hu$^{5}$ \quad Zhizhong Su$^{2}$\\[4pt]
$^{1}$UC San Diego \quad $^{2}$Horizon Robotics \quad $^{3}$Nanjing University \quad $^{4}$HKUST \quad $^{5}$Zhejiang University\\[9pt]
{\url{https://xmz111.github.io/Dianjing/}}
}

\begin{document}
\maketitle
{\renewcommand{\thefootnote}{\fnsymbol{footnote}}\footnotetext[2]{Project leader.}}
\input{sec/0_abstract}

\input{sec/1_intro}

\input{sec/2_related}

\input{sec/3_method}

\input{sec/4_exp}

{\small
\bibliographystyle{ieeenat_fullname}
\bibliography{main}
}

\end{document}

%% file: sec/0_abstract.tex
\begin{abstract}
Modern video generators excel at producing visually plausible futures guided by text and image prompts. Yet their emphasis on semantic plausibility does not guarantee fidelity to the physical consequences of specific actions. Action-conditioned world models (ACWMs), in contrast, aim to simulate how the world responds to robot actions, enabling scalable policy evaluation and improvement without costly real-world rollouts. \textbf{The challenge is not merely to generate a plausible future, but to predict what follows from a specific physical action.} As video priors grow stronger, we ask how little adaptation is truly needed to bridge this gap. We introduce \textbf{Dianjing}\footnote{Dianjing (\begin{CJK*}{UTF8}{gbsn}点睛\end{CJK*}, \textit{di\v{a}n j\={\i}ng}) comes from the Chinese idiom \mbox{\begin{CJK*}{UTF8}{gbsn}画龙点睛\end{CJK*}} (\textit{hu\`a l\'ong di\v{a}n j\={\i}ng}), ``dotting the eyes of a painted dragon'': a single finishing touch that brings an already complete painting to life.}, a minimal recipe that preserves and unlocks the capabilities of strong video priors through unified physical action conditioning, training concentrated at the high noise levels where motion and layout, and thus the consequences of actions, are decided, and native history conditioning. Built upon MiniMax H3, a 33B video foundation model, Dianjing freezes the pretrained backbone and trains only 0.34\% of its parameters using approximately 65 hours of robot video across two embodiments, without additional robot-specific pretraining or camera calibration. Controlled ablations show that where training is allocated across noise levels matters more than how actions enter the network. On WorldOracle, which evaluates predictions against paired real-world executions, Dianjing achieves the best scores on four of the five evaluation tracks and ranks second on AgiBot-Long. Our findings suggest that strong video priors already bring us most of the way to world modeling; lightweight physical action grounding can unlock their potential for real-world prediction.
\end{abstract}

%% file: sec/1_intro.tex

\begin{figure*}[t]
    \centering
    \includegraphics[width=\textwidth]{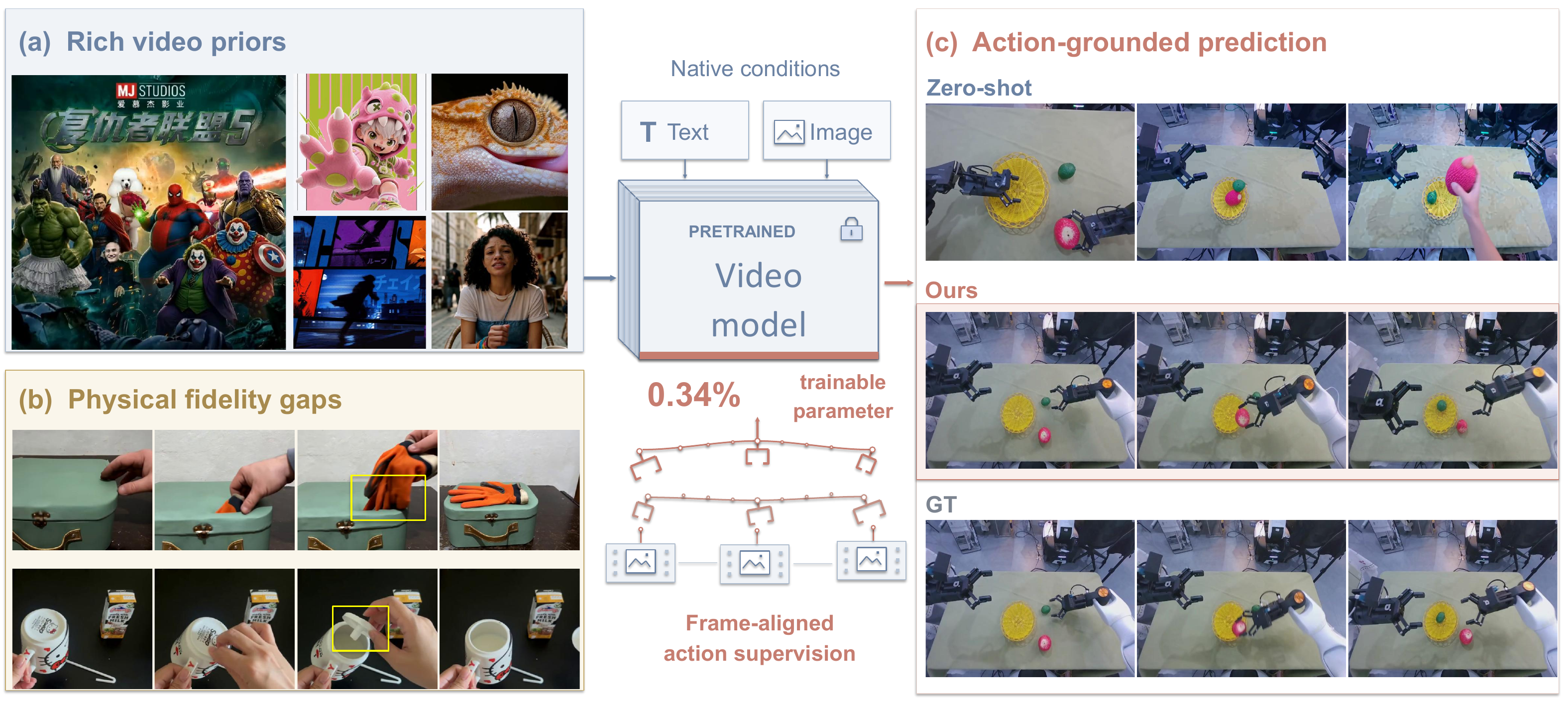}
    \caption{\textbf{Most of the way there: from rich video priors to action-grounded prediction.} (a) Modern video generators exhibit rich visual and semantic capabilities acquired through large-scale pretraining. (b) Yet visually plausible interactions can still exhibit physical inconsistencies and do not guarantee faithful prediction of specific executions. (c) The zero-shot TI2V prior generates a visually plausible but action-inconsistent future, whereas \textbf{Dianjing} grounds its predictions in temporally aligned robot actions with only 0.34\% trainable parameters, improving the episode-level SV from 31.99 to 76.74 and producing a prediction more faithful to the paired real-world execution.
}
    \label{fig:teaser}
\end{figure*}

\section{Introduction}
\label{sec:intro}

Recent advances in video generation have produced increasingly powerful models capable of synthesizing diverse scenes, realistic appearance, and complex motion and interactions~\cite{blattmann2023svd,wan2025,agarwal2025cosmos,minimax2026h3}. Trained on large-scale video data, these models acquire rich visual, semantic, and motion priors that are difficult to learn from limited robot demonstrations alone. Their growing capabilities suggest an exciting possibility: much of the generative foundation required for physical world modeling may already exist in pretrained video generators. As illustrated in Fig.~\ref{fig:teaser}(a), strong video priors offer a rich foundation upon which action-conditioned world models (ACWMs) can be built.

However, \textbf{visual plausibility does not guarantee physical fidelity}. General-purpose video generators are typically pretrained with generative objectives conditioned on text descriptions or images, rather than temporally aligned physical actions~\cite{blattmann2023svd,kong2024hunyuanvideo,wan2025}.
While this paradigm provides rich visual and motion priors, it does not explicitly ground predictions in the consequences of specific robot actions.
Even visually convincing interactions can exhibit physical inconsistencies
~\cite{li2025worldmodelbench,zhou2026paibench}, as illustrated in Fig.~1(b). More importantly, a plausible future may not correspond to the consequences of a specific action: the same instruction to grasp an object can lead to success, missed contact, or no object movement, depending on the actual execution. This distinction is fundamental to ACWMs, which must predict how the world responds to robot actions rather than merely generate plausible futures. Faithful world prediction holds the promise of enabling scalable robot policy evaluation, planning, and improvement while reducing reliance on costly real-world rollouts~\cite{yang2024unisim,guo2026ctrlworld,qiu2026gesim2}.

Interestingly, video generation and action-conditioned world modeling offer complementary strengths. From the perspective of ACWMs, pretrained video generators provide rich visual and motion knowledge acquired from large-scale video pretraining. From the perspective of video generation, robot executions offer a valuable form of supervision: temporally aligned physical actions naturally recorded alongside visual observations. Unlike high-level semantic descriptions, these commands specify concrete physical interventions, while the corresponding videos reveal their observable consequences. 
Large-scale robot datasets increasingly provide aligned
visual observations and physical actions across diverse
tasks and embodiments~\cite{oneill2024openx,agibotworld2025}.
Such action--observation pairs provide physically grounded supervision without requiring manually annotated object states, contact dynamics, or scene geometry. \textbf{Video priors provide the generative foundation, while physical actions provide the grounding for execution prediction.} The challenge is to connect these complementary strengths without relearning the visual world from scratch.

Existing approaches have explored diverse ways to bridge this gap, including large-scale robot-specific training, rendered action representations, geometric conditioning, architectural restructuring, specialized memory mechanisms, and lightweight action adaptation~\cite{qiu2026gesim2,gao2026dreamdojo,guo2026ctrlworld,rigter2025avid,chen2026h3world}. These designs address different requirements of world simulation, but also introduce varying degrees of additional training, conditioning, and architectural complexity. Meanwhile, video foundation models continue to grow in scale and capability, inviting us to reconsider how much specialized machinery is truly necessary. \textbf{As video priors grow stronger, the question is not how much more we can add, but how little we actually need to change.} We therefore ask: 
\emph{How little adaptation does it take to turn a strong video prior into a faithful action-conditioned world model?}

\begin{table*}[h]
\centering
\caption{What action-conditioned world models add on top of their video backbone. \textcolor{ours}{\textbf{Ours}} preserves the pretrained backbone and its native conditioning mechanisms. Lightweight action projections inject physical actions through AdaLN modulation, while long-horizon rollouts reuse native reference conditioning without dedicated memory modules.}
\label{tab:peers}
\small
\setlength{\tabcolsep}{3.5pt}
\renewcommand{\arraystretch}{1.12}
\begin{tabular}{lcccccc}
\toprule
Method & Full FT & Action interface & Calib. & History & Reward / judge & Stages \\
\midrule
GE-Sim 2.0~\cite{qiu2026gesim2}            & \yes & rendered map   & \yes & memory, state expert & \yes & 3 \\
EnerVerse-AC~\cite{jiang2025enerverseac}   & \yes & rendered map   & \yes & causal, memory       & \no  & 1 \\
DreamTrue~\cite{li2026dreamtrue}           & \no  & rendered map   & \yes & --                   & \yes & 2 \\
GigaWorld-1~\cite{gigaai2026gigaworld1}    & \no  & rendered map   & \yes & memory               & \no  & 4 \\
UnifoLM-WMA-0~\cite{unitree2025unifolmwma} & \yes & action head    & \no  & state tokens         & \no  & 3 \\
DreamDojo~\cite{gao2026dreamdojo}          & \yes & latent action  & \no  & AR distillation      & \no  & 3 \\
Ctrl-World~\cite{guo2026ctrlworld}         & \yes & action encoder & \no  & multi-view memory    & \no  & 1 \\
\midrule
\rowcolor{oursbg}
\oursb{Ours} & \no & \oursb{native} & \no & \oursb{native} & \no & \oursb{1} \\
\bottomrule
\end{tabular}
\end{table*}

We introduce \textbf{Dianjing}, a lightweight adaptation recipe guided by a simple philosophy: \textbf{preserve the prior and unlock its capabilities through physical action grounding}. Named after the Chinese idiom in which a painted dragon comes to life with a few finishing touches, Dianjing aims to unlock an already capable video prior rather than rebuild it. Our recipe follows three simple principles: specifying temporally aligned physical actions through a shared end-effector representation across robot embodiments, concentrating adaptation training on noise levels that most effectively improve action-conditioned prediction, and reusing the prior's native reference mechanism to carry visual history across generation chunks. We instantiate Dianjing on MiniMax H3~\cite{minimax2026h3}, a 33B video foundation model, freezing the pretrained backbone and training only approximately 0.34\% of its parameters using about 65 hours of robot video from two embodiments. This requires no additional robot-specific pretraining, camera calibration, rendered action maps, or specialized memory modules.

We investigate this approach through controlled adaptation studies and evaluate its predictions on WorldOracle~\cite{worldoracle2026}, which measures agreement with paired real-world robot executions. As illustrated in Fig.~\ref{fig:teaser}(c), minimal adaptation substantially improves execution prediction over the original video prior. Our experiments reveal that visual-only adaptation is insufficient for reliable action-conditioned prediction, while explicit physical action conditioning brings substantial gains. More elaborate action interfaces provide limited additional improvements, whereas reallocating training effort across noise levels improves validation performance by approximately 7 points at matched epochs. Despite its minimal adaptation, Dianjing achieves competitive performance against dedicated robot world models, ranking first on four of the five evaluation tracks, including 73.48 on AgiBot-Core, 70.71 on YAM-Core, and 64.25 on Policy. Although long-horizon consistency remains challenging, these results suggest that strong video priors already provide much of the foundation for action-conditioned world modeling, and that surprisingly little specialized adaptation is needed to achieve competitive real-world execution prediction.

\noindent\textbf{Our contributions are summarized as follows:}
\begin{itemize}
    \item We systematically investigate what it takes to turn a strong video prior into an ACWM, revealing the roles of pretrained generative knowledge, physical action grounding, and effective adaptation through controlled studies.
    \item We introduce \textbf{Dianjing}, a minimal adaptation recipe that grounds video priors in physical actions through a shared end-effector space, high-noise-focused training, and native reference conditioning, while preserving their pretrained backbone without architectural reconstruction.
    \item With only 0.34\% trainable parameters and about 65 hours of robot video across two embodiments, Dianjing achieves state-of-the-art scores on four of five WorldOracle tracks and ranks second on AgiBot-Long, demonstrating competitive real-world execution prediction.
\end{itemize}

%% file: sec/2_related.tex
\section{Related Work}
\label{sec:related}

\subsection{Action-Conditioned World Models}
\label{sec:rw_acwm}
An action-conditioned world model (ACWM) predicts how the world evolves under an agent's actions, an idea rooted in model-based control~\cite{ha2018worldmodels,hafner2020dreamer} and first realized in pixel space by action-conditional video prediction~\cite{oh2015action,finn2016unsupervised}. Such models let agents be planned, trained, and evaluated in imagination rather than in the real world: robots plan through predicted frames~\cite{finn2017foresight}, policies trained in learned simulators transfer to real robots~\cite{yang2024unisim}, and recent ACWMs evaluate and improve robot policies and generate training data~\cite{guo2026ctrlworld,gao2026dreamdojo,qiu2026gesim2,bwm2026}. Their evaluation has accordingly shifted from video quality and plausibility~\cite{huang2024vbench,duan2025worldscore,li2025worldmodelbench,yue2025ewmbench,zhou2026paibench} to functional utility and agreement with real executions~\cite{shang2026worldarena,shang2026worldarena2,zhang2026worldinworld,gigaai2026gigaworld1,chen2026worldecho,worldoracle2026}. 
Related efforts have explored learning interactive environments
from unlabeled videos and integrating generative world models
with robot policy learning, evaluation, and simulation
~\cite{bruce2024genie,liao2025genie}.

\subsection{Video Generation Priors}
\label{sec:rw_priors}
Video generation has advanced from early video diffusion models~\cite{ho2022vdm} to latent models trained on curated internet-scale data~\cite{blattmann2023svd}, and then to models framed as world simulators~\cite{brooks2024sora}. Open models now reach billions of parameters~\cite{yang2025cogvideox,kong2024hunyuanvideo,wan2025}, and world foundation models built for physical AI~\cite{agarwal2025cosmos,ali2025cosmospredict25,agarwal2026cosmos3} generate 3D-consistent, physically plausible video~\cite{agarwal2025cosmos}. The latest open systems reach 33B parameters in a single-stream transformer that jointly reads text, images, video, and audio~\cite{minimax2026h3}. These priors increasingly serve as the starting point for ACWMs~\cite{guo2026ctrlworld,gao2026dreamdojo,bwm2026}, yet a stronger prior does not automatically yield a better world model: transferable physical priors matter more than raw scale~\cite{gigaai2026gigaworld1}, and post-training on action--observation data can matter more than upgrading the generator~\cite{zhang2026worldinworld}.

\begin{figure*}[h]
    \centering
    \includegraphics[width=.99\textwidth]{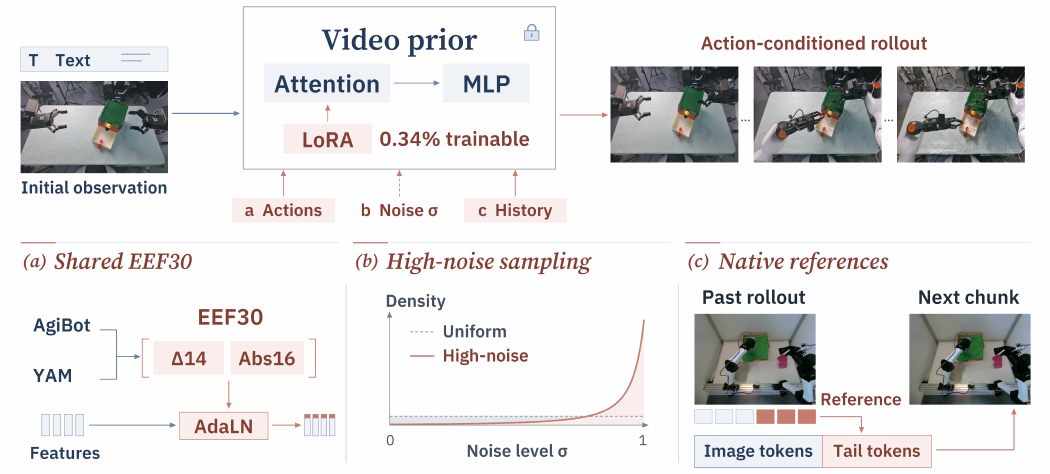}
    \caption{\textbf{Dianjing: minimal adaptation of a strong video prior for action-conditioned world modeling.} Starting from an initial observation and native text conditioning, we retain the frozen video prior and introduce only 0.34\% trainable parameters for action-conditioned rollout. (a) \textbf{Shared EEF30:} robot commands across embodiments are converted to a common, temporally aligned end-effector representation and injected through additive AdaLN residuals. (b) \textbf{High-noise sampling:} training emphasizes high noise levels to improve action-conditioned adaptation. (c) \textbf{Native references:} the clean latent tail of the generated past is passed to the next chunk through the pretrained reference pathway, without a specialized memory module.}
    \label{fig:method}
\end{figure*}

\subsection{Adapting Video Priors into World Models}
\label{sec:rw_adapt}
Pretrained video generators lack frame-level action input and causal rollout~\cite{he2025dws,huang2026vid2world}. Language offers one way in: H3-World~\cite{chen2026h3world} adapts H3 with a small LoRA by rewriting keyboard commands as text on its native text pathway, which suits simple, discrete game actions. Robot actions, however, are continuous and low-level, such as joint angles and end-effector poses, and language cannot express them precisely~\cite{rigter2025avid,zhu2024irasim}. Robot ACWMs therefore build around the generator (Table~\ref{tab:peers}), with designs well suited to the weaker priors they started from.
Full fine-tuning injects robot dynamics into priors that lacked them~\cite{qiu2026gesim2,jiang2025enerverseac,gao2026dreamdojo,guo2026ctrlworld}, but it can also erode what the prior already knows.
Rendered pose maps make actions easy to follow~\cite{gigaai2026gigaworld1,qiu2026gesim2,jiang2025enerverseac,li2026dreamtrue}, yet require calibrating every camera setup and rendering every frame; learned action encoders and heads avoid calibration but add modules trained from scratch~\cite{guo2026ctrlworld,unitree2025unifolmwma,gao2026dreamdojo}.
Memory and causal modules enable long rollouts but change how the backbone attends across time~\cite{huang2026vid2world,guo2026ctrlworld,gigaai2026gigaworld1}, while reward post-training, multi-stage recipes, and thousands of hours of data~\cite{li2026dreamtrue,qiu2026gesim2,gao2026dreamdojo} further raise the cost of every new embodiment.
These investments made sense when the prior itself was the bottleneck, and all were developed on priors of at most 14B parameters. With priors such as H3, which already produce plausible scenes but miss the physical consequences of specific actions (Sec.~\ref{sec:rw_priors}), the bottleneck shifts to grounding continuous actions. We therefore ask: what is the minimal adaptation that lets a strong prior follow physical actions, and which part of training actually unlocks it?

%% file: sec/3_method.tex
\section{Method}
\label{sec:method}

\subsection{Overview}
\label{sec:method_overview}

Our goal is to adapt a strong pretrained video generator into an action-conditioned world model (ACWM) with minimal changes to its existing architecture. Given an initial observation $I_0$, optional text condition $c$, and a sequence of robot actions $A_{1:T}$, the model predicts a future video $V_{1:T}$ that reflects the consequences of the specified actions:
\begin{equation}
    V_{1:T} \sim p_{\theta,\phi}(V_{1:T}\mid I_0,c,A_{1:T}),
    \label{eq:acwm}
\end{equation}
where $\theta$ denotes the frozen pretrained parameters and $\phi$ denotes a small set of trainable adaptation parameters. Rather than relearning visual appearance and motion dynamics from limited robot data, we preserve the pretrained generator and focus adaptation on connecting physical actions to their observable consequences.

As illustrated in Fig.~\ref{fig:method}, \textbf{Dianjing} follows three simple design choices: (a) specifying robot actions through a shared end-effector representation and injecting them into the existing modulation pathway, (b) concentrating training on noise levels that are more effective for action-conditioned learning, and (c) reusing the generator's native reference mechanism to carry visual history across rollout chunks. We instantiate Dianjing on MiniMax H3~\cite{minimax2026h3}, a 33B video foundation model. 
All pretrained backbone weights remain frozen, while
rank-32 LoRA and lightweight action projections through
AdaLN modulation~\cite{hu2022lora,peebles2023dit}
introduce approximately 114M trainable parameters,
accounting for only 0.34\% of the model.

\subsection{Physical Action Grounding}
\label{sec:method_action}

\noindent\textbf{Shared action representation.} Different robot embodiments expose different action spaces, making direct conditioning on raw commands difficult to share across datasets. We therefore adopt a unified 30-dimensional end-effector representation, termed EEF30. For AgiBot G2 and YAM, we transform their original 52-dimensional and 14-dimensional commands, respectively, into bilateral end-effector poses and gripper commands in the robot base coordinate frame using the corresponding robot kinematics. Importantly, we use commanded actions rather than measured future states, avoiding information leakage from the target execution.

Each EEF30 action combines relative and absolute information:
\begin{equation}
    \mathbf{a}_t =
    \left[\Delta\mathbf{a}_t,\mathbf{a}^{\mathrm{abs}}_t\right]
    \in\mathbb{R}^{30},
    \qquad
    \Delta\mathbf{a}_t\in\mathbb{R}^{14},
    \quad
    \mathbf{a}^{\mathrm{abs}}_t\in\mathbb{R}^{16}.
    \label{eq:eef30}
\end{equation}
The relative component contains changes in position, Euler orientation, and gripper commands for both arms, while the absolute component contains positions, quaternion orientations, and gripper commands. This representation provides both local motion information and global spatial context. We resample the action sequence to match the temporal resolution of the video latents, yielding $\mathbf{A}\in\mathbb{R}^{37\times30}$ for each 124-frame training window. Action values are normalized using training-set statistics.

\noindent\textbf{Lightweight action injection.} We inject EEF30 actions through the pretrained model's existing adaptive layer normalization (AdaLN) pathway. For each transformer block $\ell$, a lightweight linear projection maps the action at latent timestep $t$ into six modulation components corresponding to the shift, scale, and gate parameters of the attention and feed-forward branches. Rather than modifying the pretrained timestep modulation, we introduce an additive action residual:
\begin{equation}
    \widetilde{\mathbf{m}}_{\ell,t}
    =
    \mathbf{m}_{\ell}(\sigma)
    +
    W_{\ell}\mathbf{a}_t,
    \qquad
    W_{\ell}\in\mathbb{R}^{6d\times30},
    \label{eq:action_injection}
\end{equation}
where $\mathbf{m}_{\ell}(\sigma)$ is the original noise-dependent modulation and $d$ is the hidden dimension. The action projections $W_{\ell}$ are zero-initialized, ensuring that the action pathway initially leaves the pretrained modulation unchanged. Action modulation is applied only to video tokens at their corresponding latent temporal positions, leaving text and reference tokens unaffected. This design introduces action conditioning without additional attention branches, rendered action images, or camera calibration, while retaining the original timestep modulation pathway.

\subsection{Training Where Events Are Decided}
\label{sec:noise}
An ACWM is judged by what happens: whether the gripper makes contact, whether the object moves, and where it ends up~\cite{worldoracle2026}. A strong prior already knows how such scenes look; what it must learn is which event a given action produces. Along the denoising trajectory, these two aspects are formed at different stages. Image diffusion models learn coarse structure and content at high and intermediate noise levels and mostly remove residual noise at low ones~\cite{choi2022p2}, and conditioning signals act mainly early in sampling~\cite{balaji2022ediffi}. Video diffusion models follow the same pattern: motion and layout are established at high noise levels, while appearance is refined at low ones~\cite{baherwani2025motion,luo2026golden}. The consequences of an action are therefore decided largely at high noise levels, whereas low-noise steps mostly render appearance details that the pretrained prior already handles well.

Prior work has studied timestep-dependent loss weighting for perceptual quality and optimization efficiency~\cite{choi2022p2,hang2023minsnr}. Our focus is instead on allocating training effort to improve action-conditioned prediction.
We therefore direct adaptation toward high noise levels. We keep the pretrained flow-matching objective~\cite{lipman2023flow} and change only how training is distributed across noise levels: the training timestep distribution is shifted toward high noise~\cite{esser2024sd3}, and the original bell-shaped loss weighting is replaced with uniform weighting over sampled timesteps. As a result, more than half of the training signal falls at noise levels $\sigma\geq0.9$, where layout and motion take shape. Timestep-aware allocation has improved motion learning in video generation~\cite{luo2026golden}; here we use it to learn the consequences of actions.

This change adds neither a new objective nor trainable parameters. We refer to it as high-noise-focused training, without assuming that action information is learned only at high noise levels. As our experiments show, it improves action-conditioned prediction more than richer action interfaces do.

\subsection{Long-Horizon Rollouts with Native References}
\label{sec:method_history}

ACWMs must predict extended sequences of robot interactions, whereas pretrained video generators typically operate on shorter clips. We generate long-horizon predictions autoregressively by dividing the action sequence into consecutive chunks, each containing 124 frames. Rather than introducing a dedicated memory module, we reuse the pretrained generator's native reference-to-video conditioning mechanism. 
Long-horizon generation has been explored through
variable-horizon diffusion, training with self-generated
contexts, and history-aware context packing
~\cite{chen2024df,huang2025selfforcing,zhang2025framepack}.

\begin{table*}[h]
\centering
\caption{\textbf{WorldOracle official test results (SV$\uparrow$).}
We compare Dianjing with certified baselines across Core,
Policy, and Long tracks using the official online leaderboard. Data denotes the reported training-data scale. Our model is listed as HWM-AC on the official leaderboard.
\textbf{Bold} and \underline{underline} indicate the best
and second-best scores in each column.}
\label{tab:worldoracle_main}
\begingroup
\small
\setlength{\tabcolsep}{12pt}
\renewcommand{\arraystretch}{1.17}
\begin{tabular}{lcccccc}
\toprule
\multirow{2}{*}{\textbf{Method}}
& \multicolumn{2}{c}{\textcolor{prior}{\textbf{Core}}}
& \textcolor{prior}{\textbf{Policy}}
& \multicolumn{2}{c}{\textcolor{prior}{\textbf{Long}}}
& \multirow{2}{*}{\textbf{Data}} \\
\cmidrule(lr){2-3}
\cmidrule(lr){4-4}
\cmidrule(lr){5-6}
& AgiBot & YAM & AgiBot & AgiBot & YAM & \\
\midrule

Vid2World~\cite{huang2026vid2world}
& 36.94 & 36.82 & 32.86 & 18.68 & 16.64
& $\sim$333\,h \\

ABot-PhysWorld~\cite{chen2026abotphysworld}
& 48.48 & 39.46 & 30.74 & 18.87 & 13.24
& $\sim$3M clips \\

IRASim~\cite{zhu2024irasim}
& 50.94 & 47.75 & 44.87 & 37.20 & 36.80
& $\sim$440k eps. \\

Ctrl-World~\cite{guo2026ctrlworld}
& 51.51 & 45.60 & 33.91 & 32.85 & 23.02
& 350\,h \\

DreamDojo~\cite{gao2026dreamdojo}
& 52.22 & 51.08 & 42.36 & 29.15 & 28.52
& 44.7k\,h \\

Cosmos 3~\cite{agarwal2026cosmos3}
& 55.03 & 55.81 & 41.84 & 28.30 & 32.71
& 61.3k\,h \\

GigaWorld-1~\cite{gigaai2026gigaworld1}
& 56.69 & 57.50 & 38.96 & 45.45 & 46.40
& 12,980\,h \\

BWM~\cite{bwm2026}
& 66.08 & 62.91 & 57.26
& 54.53 & \underline{54.73}
& $>$2k videos \\

OpenDW~\cite{dexmal2026opendw}
& 66.48 & 61.82 & 54.34 & 48.39 & 46.55
& Multi-source \\

GE-Sim 2.0~\cite{qiu2026gesim2}
& \underline{73.20} & \underline{62.97}
& \underline{62.61} & \textbf{62.57} & 49.34
& $>$1k\,h \\

\midrule
\rowcolor{oursbg}
\textcolor{ours}{\textbf{Dianjing (Ours)}}
& \textcolor{ours}{\textbf{73.48}}
& \textcolor{ours}{\textbf{70.71}}
& \textcolor{ours}{\textbf{64.25}}
& \textcolor{ours}{\underline{56.43}}
& \textcolor{ours}{\textbf{54.76}}
& \textcolor{ours}{\textbf{65\,h}} \\

\bottomrule
\end{tabular}

\vspace{3pt}

\endgroup
\end{table*}

After generating chunk $k$, we retain its final three clean video latents before decoding. These latents are supplied as reference tokens for the next chunk, together with the initial image reference. The next chunk is generated under its corresponding action sequence:
\begin{equation}
    V^{(k+1)}
    \sim
    p_{\theta,\phi}
    \left(
    V^{(k+1)}
    \mid I_0,c,A^{(k+1)},R^{(k)}
    \right),
    \label{eq:long_rollout}
\end{equation}
where $R^{(k)}$ denotes the clean latent tail of the previously generated chunk. Unlike RGB-frame handoff, this strategy transfers the generated latents directly without repeated decoding and re-encoding. The reference latents enter through the native reference pathway and remain separate from the action-modulated target video tokens.

This approach extends the pretrained generator to chunk-wise action-conditioned rollouts without additional memory parameters or modifications to its backbone. While native latent references provide a simple, parameter-free mechanism for temporal continuation, maintaining physical consistency over long horizons remains challenging for chunk-wise generation, as examined in our experiments.

%% file: sec/4_exp.tex
\section{Experiments}
\label{sec:exp}

\subsection{Evaluation Setup}
\label{sec:exp_setup}

\noindent\textbf{Benchmark and metrics.} We evaluate Dianjing on WorldOracle~\cite{worldoracle2026}, which compares action-conditioned video predictions against paired real-world robot executions. We report the single-view score (SV), which averages physical state correctness (PSC), key interaction area fidelity (KIAF), and interaction spatial alignment (ISA). The five evaluation panels span two embodiments (AgiBot G2 and YAM), human-operated Core and Long rollouts, and AgiBot policy-action rollouts. For ablations, we use the fixed 47-episode validation suite and report the equal-weight average of its five panels; the validation PSC is computed with our fixed internal evaluator, whereas the test scores are returned by the official service.

\paragraph{Training data and implementation.}

We train on the WorldOracle training suite~\cite{worldoracle2026}, using 1,779 AgiBot G2 episodes and a fixed selection of 2,089 YAM episodes from ABC-130K~\cite{allshire2026abc}, totaling approximately 65 hours of robot video. The original WorldOracle study similarly adapts robot-oriented baselines to the target embodiments using the corresponding training data, but several of these methods are initialized from checkpoints that have already undergone robot-domain training~\cite{worldoracle2026}. In contrast, Dianjing directly adapts a general-purpose video foundation model using this 65-hour suite, without an additional robot-specific pretraining stage. Actions are converted to EEF30 from the commanded joints via forward kinematics; no camera calibration and no validation or test data are used. All pretrained H3 weights remain frozen throughout training; we train rank-32 LoRA on the DiT attention projections together with the zero-initialized AdaLN action projections (114M parameters, 0.34\%), using AdamW with a learning rate of $10^{-4}$ and a global batch size of 32 on 32 NVIDIA H20 GPUs. The main model is trained for 8 epochs.

\subsection{Main Results}
\label{sec:main_results}

\noindent\textbf{Competitive performance with minimal adaptation.} Table~\ref{tab:worldoracle_main} compares Dianjing (submitted as HWM-AC by Horizon Robotics) with ten certified baselines on the official WorldOracle leaderboard. Despite its minimal adaptation, Dianjing achieves the highest reported scores on four of the five evaluation tracks: AgiBot-Core (73.48), YAM-Core (70.71), Policy (64.25), and YAM-Long (54.76), and ranks second on AgiBot-Long (56.43). Dianjing also substantially outperforms existing general video-prior-based approaches such as Cosmos 3, DreamDojo, and Ctrl-World across all five tracks. These results demonstrate that a strong pretrained video generator can achieve highly competitive real-world execution prediction across different evaluation settings without extensive robot-specific architectural reconstruction.

\begin{table*}[h]
\centering
\caption{\textbf{Step-by-step ablation on WorldOracle
validation (SV$\uparrow$).}
Each row progressively adds one component to the
pretrained video prior. PSC, KIAF, and ISA are averaged
over the five evaluation panels. Avg is the equal-weight
mean of the five panel scores.
\textbf{Bold} indicates the best score in each column.}
\label{tab:ablation}
\begingroup
\small
\setlength{\tabcolsep}{4.5pt}
\renewcommand{\arraystretch}{1.17}
\begin{tabular}{lccccccccc}
\toprule
\multirow{2}{*}{\textbf{Method}}
& \multicolumn{2}{c}{\textcolor{prior}{\textbf{Core}}}
& \textcolor{prior}{\textbf{Policy}}
& \multicolumn{2}{c}{\textcolor{prior}{\textbf{Long}}}
& \multirow{2}{*}{PSC}
& \multirow{2}{*}{KIAF}
& \multirow{2}{*}{ISA}
& \multirow{2}{*}{\textbf{Avg}} \\
\cmidrule(lr){2-3}
\cmidrule(lr){4-4}
\cmidrule(lr){5-6}
& AgiBot & YAM & AgiBot & AgiBot & YAM
& & & & \\
\midrule

H3 (zero-shot)
& 26.37 & 30.78 & 30.41 & 16.19 & 18.69
& 38.41 & 15.50 & 19.56 & 24.49 \\

+ Visual-only
& 45.08 & 49.41 & 37.99 & 27.07 & 33.70
& 47.89 & 29.90 & 38.16 & 38.65 \\

+ Action (EEF30 + AdaLN)
& 61.59 & 60.23 & 64.37 & 37.23 & 40.01
& 60.84 & 38.35 & 58.87 & 52.69 \\

+ High-noise training
& 69.95 & \textbf{67.91} & 65.87
& 49.88 & 52.11
& 69.67 & 44.44 & 69.32
& 61.15 \\

\midrule
\rowcolor{oursbg}
\oursb{+ Native references (Ours)}
& \oursb{71.72}
& \oursb{66.62}
& \oursb{66.78}
& \oursb{54.66}
& \oursb{56.86}
& \oursb{70.15}
& \oursb{46.36}
& \oursb{73.47}
& \oursb{63.33} \\

\bottomrule
\end{tabular}

\endgroup
\end{table*}

\noindent\textbf{Reusing large-scale priors instead of robot-specific pretraining.} The performance is particularly notable given the different training strategies of existing world models. GE-Sim 2.0~\cite{qiu2026gesim2} is retrained on thousands of hours of real-world robot data, with a backbone pretrained on AgiBot G1, an embodiment closely related to the AgiBot G2 evaluation platform~\cite{worldoracle2026}. DreamDojo~\cite{gao2026dreamdojo} introduces 44,000 hours of human egocentric video pretraining, while GigaWorld-1~\cite{gigaai2026gigaworld1} uses a heterogeneous corpus of approximately 13,000 hours. In contrast, Dianjing directly adapts an existing general-purpose video prior using only approximately 65 hours of action-labeled robot video across two embodiments, without additional robot-specific pretraining or camera calibration. Its strong performance, particularly on YAM-Core, where it leads the second-best method by 7.7 points, suggests that much of the generative knowledge needed for robot world prediction can be reused from large-scale video pretraining rather than learned again from robot-specific data.

\begin{figure*}[h]
\centering
\includegraphics[width=\textwidth]{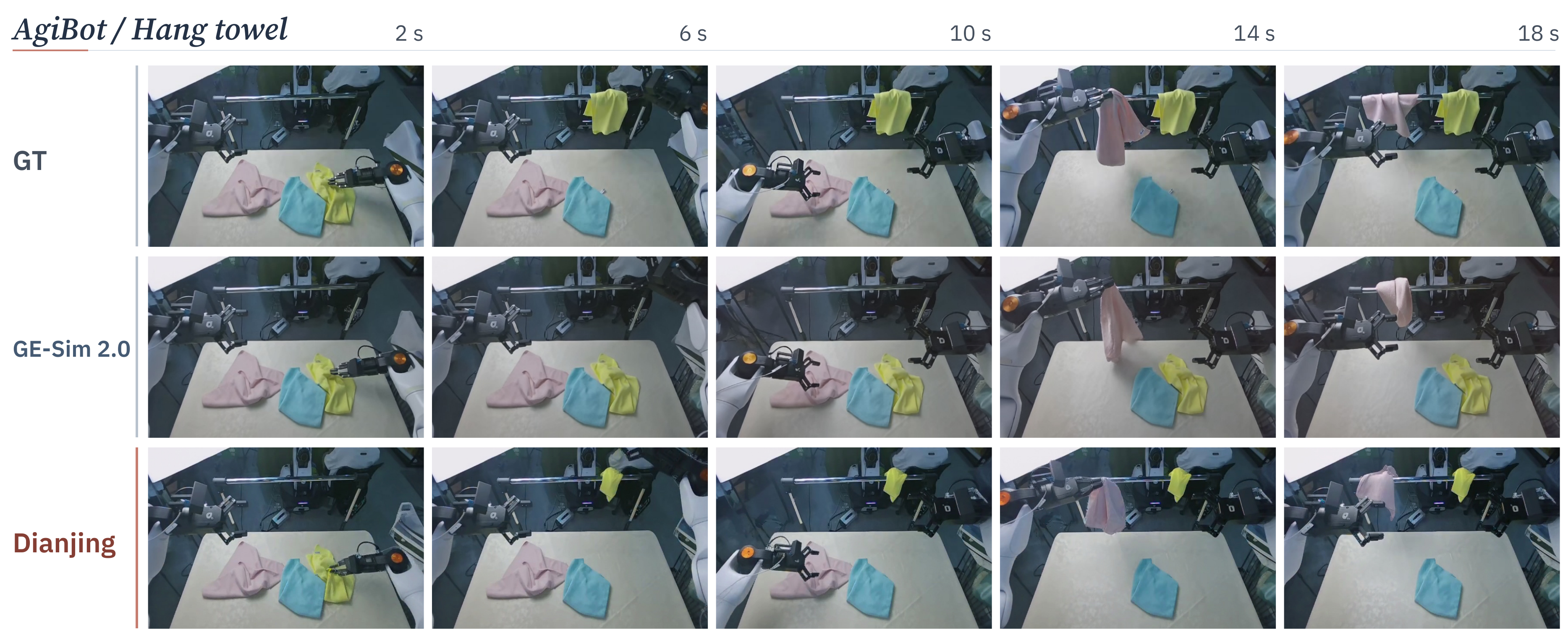}
\caption{\textbf{Qualitative comparison} on a WorldOracle validation episode (AgiBot Core, hang towel), in which the robot hangs the yellow towel and then the pink one. GE-Sim 2.0 is run with its released WorldOracle checkpoint and inference script. It moves the arms as commanded but leaves the yellow towel on the table; Dianjing reproduces both hanging events.}
\label{fig:qual}
\end{figure*}

\subsection{Ablation Studies}
\label{sec:ablation}

\noindent\textbf{From visual adaptation to physical action grounding.}
Table~\ref{tab:ablation} progressively examines how a
strong video prior can be adapted into an action-conditioned
world model. The zero-shot H3 prior achieves an average
SV of 24.49. Fine-tuning on robot videos without action
conditioning improves the score to 38.65, indicating
that visual adaptation alone provides substantial gains.
Introducing EEF30 action conditioning through AdaLN
further raises the score to 52.69. Notably, the Policy
score improves from 37.99 to 64.37, while ISA increases
by 20.71 points. These results highlight the importance of grounding the pretrained video prior in explicit, temporally aligned physical actions rather than relying on visual adaptation alone.

\noindent\textbf{Effective adaptation through noise allocation.}
High-noise-focused training further improves the average
SV from 52.69 to 61.15, with gains across all five
evaluation panels. The improvements are particularly
pronounced on the Long tracks, where AgiBot-Long and
YAM-Long increase by 12.65 and 12.10 points,
respectively. PSC, KIAF, and ISA also improve by
8.83, 6.09, and 10.45 points. Importantly, these gains
require no additional action-conditioning modules,
but only a redistribution of the existing training
signal across noise levels. 

\noindent\textbf{Long-horizon prediction.} Dianjing achieves 56.43 on AgiBot-Long and 54.76 on YAM-Long, ranking second and first by reported score, respectively. Despite its competitive performance against existing specialized world models, its scores still drop substantially from the corresponding Core tracks. In particular, GE-Sim 2.0 maintains a clear advantage on AgiBot-Long, while Dianjing achieves a marginally higher score than BWM on YAM-Long. These results indicate that lightweight adaptation is effective for action-conditioned prediction, but maintaining reliable physical and visual consistency over extended autoregressive rollouts remains an important challenge for future research.

\subsection{Porting Prior Paradigms to a Strong Prior}
\label{sec:paradigms}
Reported scores of prior ACWMs reflect differences in backbone, data, and training together. To compare design choices directly, we re-implement representative paradigms in our H3 pipeline and add them one at a time, with the same data and training budget (Table~\ref{tab:paradigms}).

\paragraph{Action interfaces.}
Action classifier-free guidance~\cite{huang2026vid2world} lowers the score by 0.4--3.6 points. Separate action tokens from a small MLP encoder, similar to Ctrl-World~\cite{guo2026ctrlworld}, help under the default noise schedule but bring no consistent gain after high-noise training. Trajectory maps concatenated with the video patches, drawn on a fixed robot-frame canvas, also bring no consistent gain over AdaLN. On a strong prior, how training is allocated across noise levels matters more than how actions enter the model.

\paragraph{Calibration and multiple embodiments.}
Pixel-aligned action maps~\cite{qiu2026gesim2} relate actions directly to image locations, which makes them easy to follow, but they depend on camera calibration for every setup. Calibration is not available in every dataset, and keeping it consistent is harder when data from several robots are combined; recovering it adds an estimation step that often relies on other models. For YAM, WorldOracle provides no calibration, and its GE-Sim 2.0 baseline fits each episode's cameras from fingertip alignment and multi-view 3D reconstruction~\cite{worldoracle2026}. We therefore leave calibrated maps aside. Since our actions are given in each robot's base frame, a single checkpoint covers AgiBot and YAM, and adding YAM required only its kinematics.

\paragraph{Long horizons.}
Longer generation windows lower the score, even though H3 can extend videos in time. Pinning the first frame as an anchor leads the model to copy it (AgiBot-Long $-$10.5), and RGB handoff lets spatial errors accumulate across chunks. Passing the clean latent tail of the previous chunk through H3's native reference pathway raises the long-horizon average by 4.8 points without new parameters. Long-horizon consistency remains our main gap, and closing it likely needs more than chunk-level latent references, such as memory that spans the whole episode.

\begin{table}[h]
\centering
\caption{Other paradigms on our backbone (validation SV$\uparrow$, five-track average). Each block starts from a base, and each row adds one change to that base, trained on the same data for the same budget (shorter than in Table~\ref{tab:ablation}). $\Delta$ is relative to the block's base.}
\label{tab:paradigms}
\small
\setlength{\tabcolsep}{4pt}
\begin{tabular}{lcc}
\toprule
Setting & Avg & $\Delta$ \\
\midrule
Base: EEF30 + AdaLN & 51.59 & -- \\
\quad + Robot-frame trajectory map~\cite{qiu2026gesim2} & 50.11 & $-$1.5 \\
\quad + Action CFG~\cite{huang2026vid2world} & 47.96 & $-$3.6 \\
\quad + 243-frame generation window & 49.07 & $-$2.5 \\
\midrule
Base: EEF30 + AdaLN + high-noise & 58.43 & -- \\
\quad + Separate action tokens~\cite{guo2026ctrlworld} & 57.98 & $-$0.4 \\
\bottomrule
\end{tabular}
\end{table}

\begin{figure*}[h]
\centering
\includegraphics[width=\textwidth]{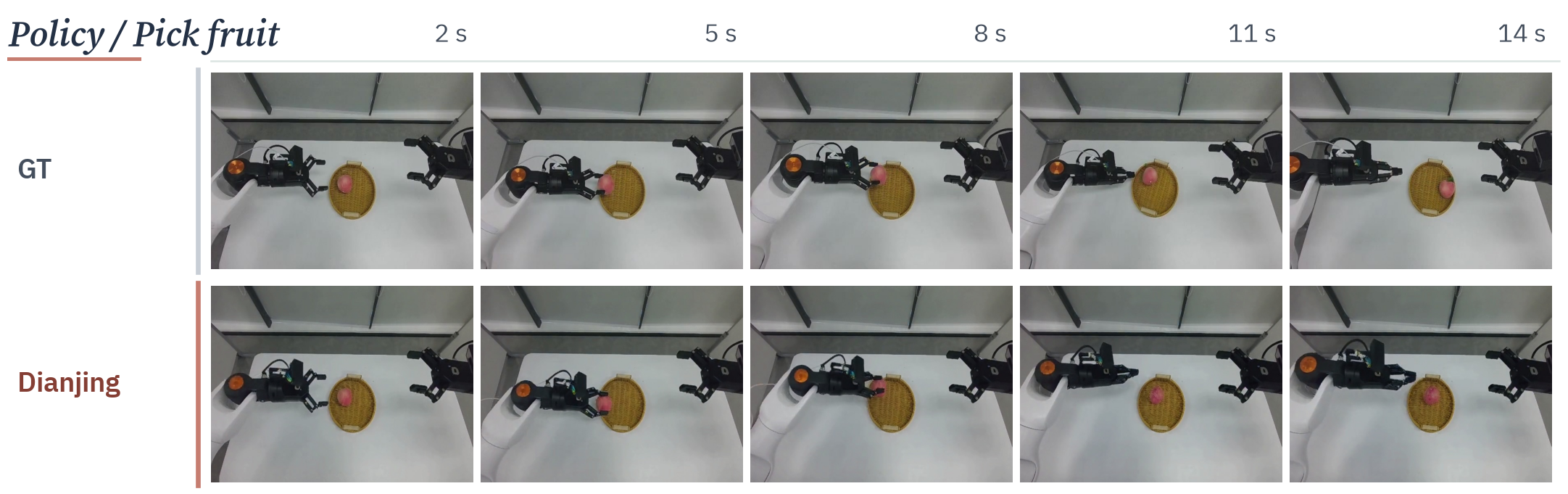}\\[4pt]
\includegraphics[width=\textwidth]{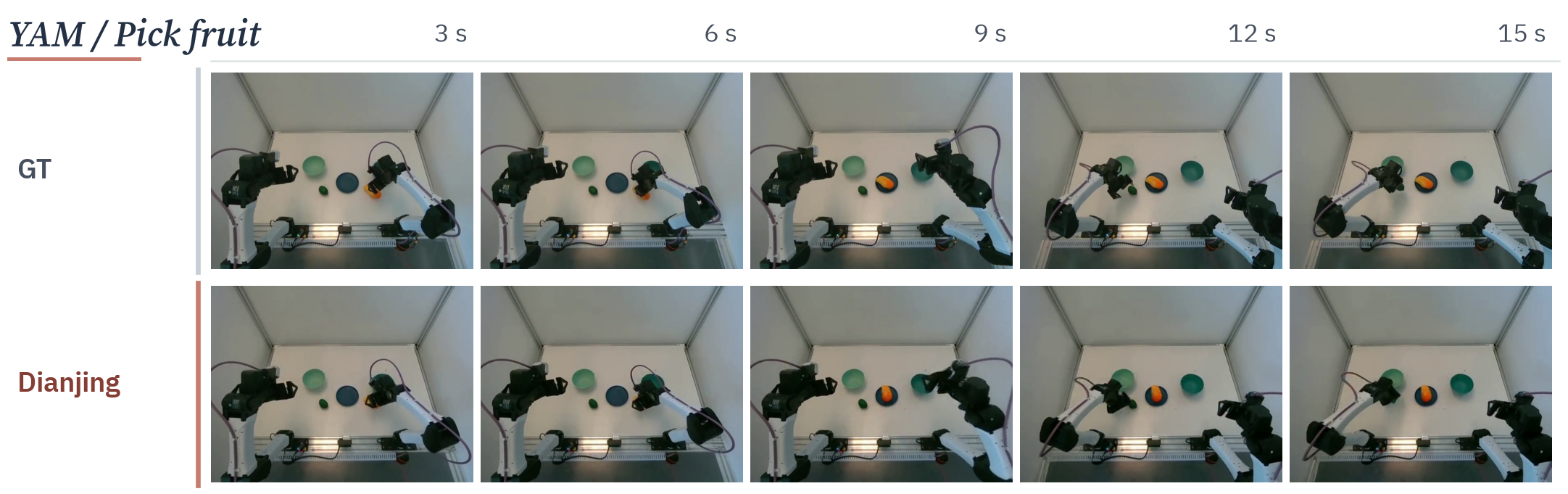}
\caption{\textbf{Additional predictions on WorldOracle validation episodes.} Top: AgiBot G2 policy rollout (pick fruit). Bottom: human-operated YAM episode (pick fruit). In each block, the top row shows the real execution and the bottom row shows Dianjing's prediction.}
\label{fig:qual_additional}
\end{figure*}

\subsection{Qualitative Results}
\label{sec:qual}

\paragraph{Event-level prediction.}
Fig.~\ref{fig:qual} compares Dianjing with GE-Sim 2.0 on a WorldOracle validation episode in which the robot hangs a yellow towel followed by a pink one. GE-Sim 2.0 follows the commanded arm motions but leaves the yellow towel on the table, whereas Dianjing reproduces both hanging events in the correct temporal order. This example illustrates the distinction between following robot motion and predicting its physical consequences on the environment. Even when the predicted arm motion is accurate, the resulting object states may differ substantially from the real execution. In particular, evaluating sequential manipulation tasks requires considering not only individual interactions but also how the resulting changes in the scene evolve over time. Such event-level differences are particularly relevant when world models are used for policy evaluation, where both the execution process and the resulting task outcomes are of interest.
\paragraph{Prediction across embodiments and action sources.}
Fig.~\ref{fig:qual_additional} presents additional predictions on AgiBot G2 and YAM. In the AgiBot policy rollout, Dianjing approximately reproduces the grasp, lift, and release of the fruit while the idle arm remains stationary. In the human-operated YAM episode, it captures the sequential interactions of both arms, placing the orange on the plate and then grasping the green fruit. These examples show that a single checkpoint can support different embodiments and action sources without camera calibration or embodiment-specific modifications.

\paragraph{Limitations.}
Although Dianjing captures the major interaction events, discrepancies remain in object positions, orientations, and fine-grained physical states, and they grow over extended rollouts as errors accumulate across chunks. Faithful robot motion alone does not guarantee accurate object states, and long-horizon consistency remains an important direction for future work.

\section{Conclusion}
We asked how little adaptation it takes to turn a strong video prior into an action-conditioned world model. Dianjing keeps the 33B prior frozen and builds on its native pathways: actions in a shared end-effector space enter through the existing AdaLN modulation, training is concentrated at high noise levels, and rollouts extend through native latent references. With 0.34\% trainable parameters and about 65 hours of robot video, Dianjing achieves the highest scores on four of the five WorldOracle tracks and ranks second on AgiBot-Long. These results suggest that strong video priors already provide much of the generative foundation needed for effective action-conditioned world prediction. Long-horizon consistency remains an important challenge, and improving it while preserving the pretrained video prior is a natural next step.